\documentclass{article}

\usepackage[preprint]{neurips_2025}

\usepackage[utf8]{inputenc} 
\usepackage[T1]{fontenc}    
\usepackage{hyperref}       
\usepackage{url}            
\usepackage{booktabs}       
\usepackage{amsfonts}       
\usepackage{nicefrac}       
\usepackage{microtype}      
\usepackage{xcolor}         

\usepackage{epsfig}
\usepackage{graphicx}
\usepackage{amsmath}
\usepackage{amssymb}
\usepackage{multirow}
\usepackage{makecell}
\usepackage{wrapfig}
\usepackage{pifont}
\usepackage{colortbl}

\title{DetailFlow: 1D Coarse-to-Fine Autoregressive Image Generation via Next-Detail Prediction}

\author{
Yiheng Liu\footnotemark[1], 
Liao Qu\footnotemark[1], 
Huichao Zhang, 
Xu Wang\footnotemark[2],
Yi Jiang, Yiming Gao, Hu Ye, \\
\textbf{Xian Li, Shuai Wang, Daniel K. Du, 
Fangmin Chen, Zehuan Yuan, Xinglong Wu} \\
ByteDance Inc. \\
\textcolor{cyan}{\href{https://github.com/ByteFlow-AI/DetailFlow}{https://github.com/ByteFlow-AI/DetailFlow}}
}

\begin{document}

\maketitle

\footnotetext[1]{\(^\star\) Equal contribution.}
\footnotetext[2]{\(^\dagger\) Project leader.}

\begin{figure}[h]
    \begin{center}
    \includegraphics[width=0.62\linewidth]{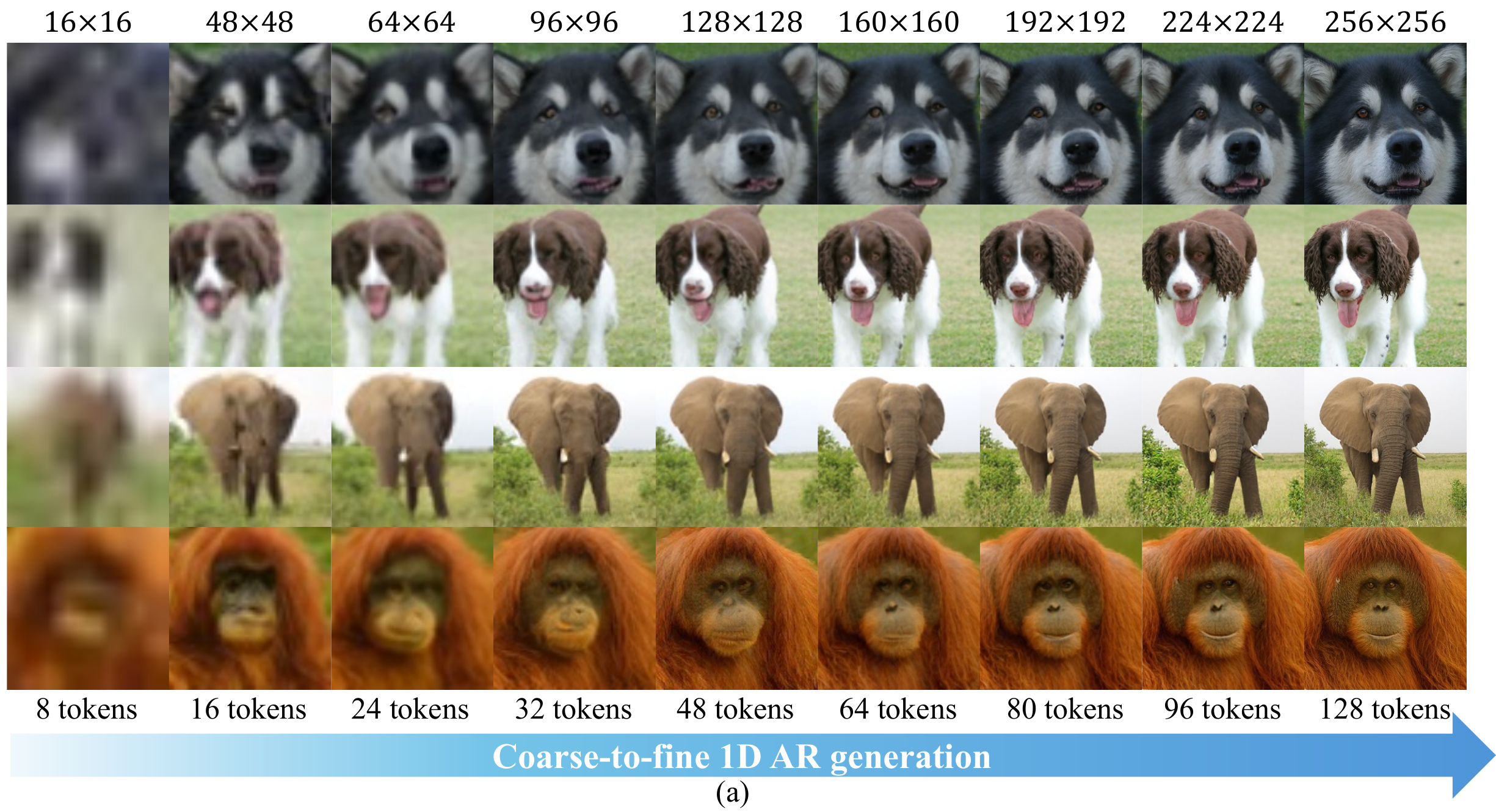}
    \includegraphics[width=0.35\linewidth]{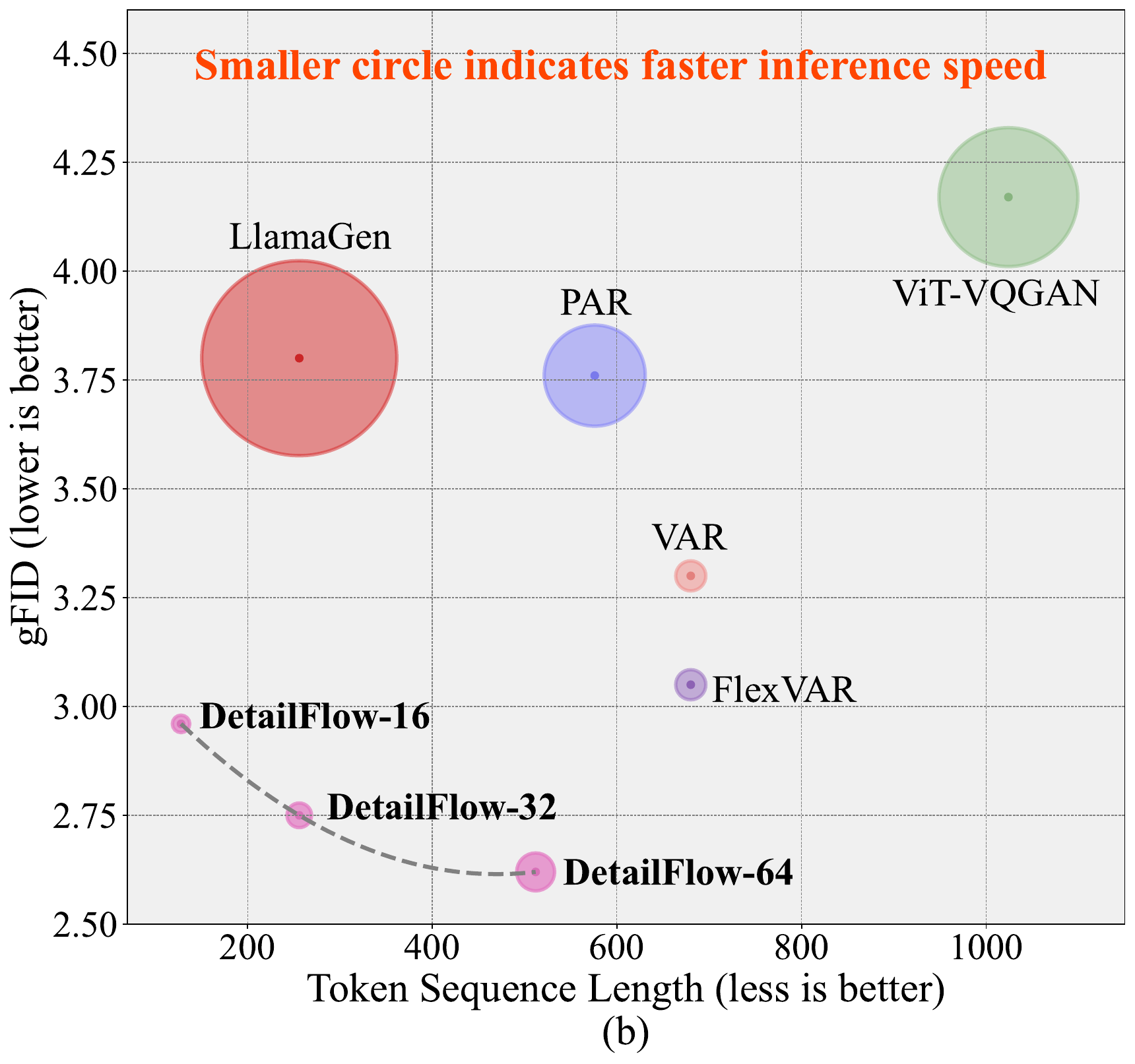}
    \end{center}
    \caption{(a) Progressive generation results from DetailFlow. Our proposed 1D tokenizer encodes tokens with an inherent semantic ordering, where each subsequent token contributes additional high-resolution information. The sequences illustrate how image resolution and inferred 1D tokens incrementally increase from left to right. (b) Comparison of our DetailFlow approach with existing methods, showing that DetailFlow achieves better image quality with fewer tokens and times.}
    \label{fig:demo:generation}
    \end{figure}

\begin{abstract}
    This paper presents \textbf{DetailFlow}, a coarse-to-fine 1D autoregressive (AR) image generation method that models images through a novel next-detail prediction strategy. By learning a resolution-aware token sequence supervised with progressively degraded images, DetailFlow enables the generation process to start from the global structure and incrementally refine details. This coarse-to-fine 1D token sequence aligns well with the autoregressive inference mechanism, providing a more natural and efficient way for the AR model to generate complex visual content. Our compact 1D AR model achieves high-quality image synthesis with significantly fewer tokens than previous approaches, i.e. VAR/VQGAN. We further propose a parallel inference mechanism with self-correction that accelerates generation speed by approximately 8× while reducing accumulation sampling error inherent in teacher-forcing supervision. On the ImageNet 256\(\times\)256 benchmark, our method achieves 2.96 gFID with 128 tokens, outperforming VAR (3.3 FID) and FlexVAR (3.05 FID), which both require 680 tokens in their AR models. Moreover, due to the significantly reduced token count and parallel inference mechanism, our method runs nearly 2× faster inference speed compared to VAR and FlexVAR. Extensive experimental results demonstrate DetailFlow's superior generation quality and efficiency compared to existing state-of-the-art methods.
\end{abstract}

\section{Introduction}

Autoregressive (AR) models like \cite{radford2018improving, radford2019language, brown2020language, chowdhery2023palm, anil2023palm, hoffmann2022training, touvron2023llama, le2023bloom} have demonstrated exceptional success in natural language processing through their scalability, flexibility, and ability to model complex sequential dependencies. Building on these strengths, researchers have extended AR modeling to image generation, creating unified frameworks \cite{team2024chameleon, wang2024emu3, xie2024show, qu2024tokenflow} for visual generation tasks. AR image generation enables structured, step-by-step synthesis, offering advantages in controllability and multimodal integration.

\begin{wrapfigure}{r}{0.56\textwidth} 
    \includegraphics[width=0.54\textwidth]{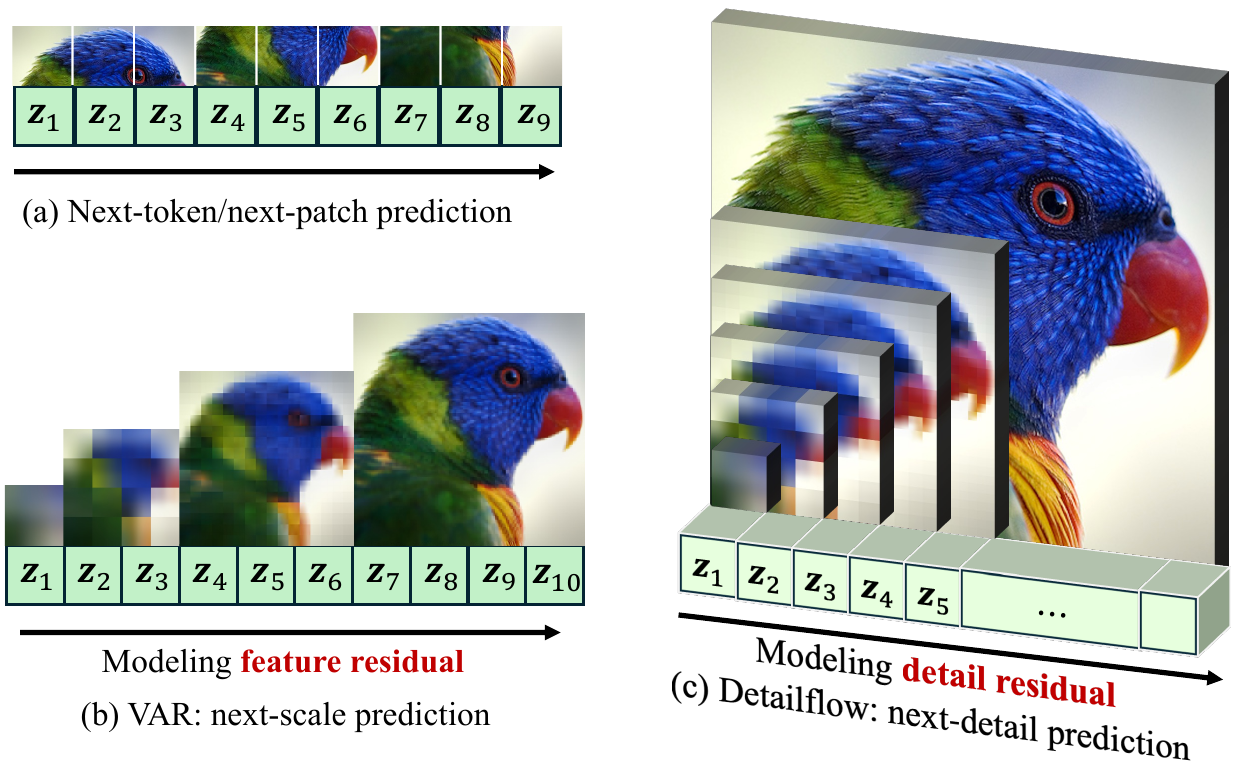} 
    \caption{Comparison of different prediction strategies in image generation. (a) Traditional 2D raster-scan next-token/next-patch prediction. (b) Next-scale prediction in VAR \cite{tian2024visual}. (c) Our proposed next-detail prediction, which predicts 1D tokens encoding fine-grained details for high-resolution image generation. }
    \label{fig:ar:next}
\end{wrapfigure}

Conventional AR image generation methods adopts raster scan approaches \cite{esser2020taming, yu2022scaling, wang2024emu3}, which flattens 2D image tokens into 1D sequences, forcing models to predict patches in a counter-intuitive order that disrupts spatial continuity. 
Recent work Visual Autoregressive Modeling (VAR) \cite{tian2024visual} adopts a next-scale prediction framework that emulates human sketching via coarse-to-fine 2D parallel generation. However, it requires extensive volumes of multi-scale tokens, particularly at high resolutions. For instance, Infinity \cite{han2024infinity} requires 10,521 tokens to synthesize a 1024×1024 image, leading to substantial computational and memory overhead during training. This bottleneck highlights a critical challenge in balancing generation quality with efficiency for high-resolution AR image synthesis.

Recent line of research \cite{yu2024image,bachmann2025flextok} attempt to address this challenge through query-based attention mechanisms that compress 2D images into adaptive 1D token sequences. These approaches remove constraints associated with fixed spatial positions, enabling adaptive compression of spatial redundancy. This significantly reduces token counts, alleviating computational overhead during generation. However, current 1D tokenization methods remain inherently constrained by resolution-specific tokenizers, limiting flexibility in generating images at arbitrary resolutions without retraining.

To address these limitations, we propose \textit{DetailFlow}, a novel coarse-to-fine 1D tokenizer explicitly establishing a semantic and resolution-dependent mapping between token sequences and image resolutions. Specifically, DetailFlow employs progressively degraded images to supervise token sequences of increasing lengths, inherently embedding a coarse-to-fine semantic ordering within the tokens. Consequently, during inference, DetailFlow autoregressively generates tokens in a coarse-to-fine manner, progressively outputing higher-resolution images with enriched visual details, as illustrated in Fig. \ref{fig:demo:generation}(a). This approach allows 256×256 images to be represented with just 128 semantically ordered tokens, significantly fewer than VAR-based methods. With similar AR model sizes, DetailFlow achieves a gFID score below 3, demonstrating superior quality and efficiency.

Furthemore, traditional VQGAN-based raster-scan methods inherently disable \textit{parallel prediction of successive tokens} due to spatial dependencies. DetailFlow's 1D learnable latent space offers greater flexibility. This enables substantial acceleration of inference speed through parallel successive token prediction (speedup proportional to the parallel token numbers).

During experimentation, we observe that sampling errors significantly degrade subsequent generation quality due to autoregressive teacher forcing training regimes. To mitigate this accumulation errors, we introduce a self-correction training strategy during tokenizer learning. Specifically, controlled quantization errors are introduced during token quantization, and subsequent tokens are trained to correct these inaccuracies. This approach fosters token sequences capable of self-correction, significantly enhancing overall generation quality and providing an effective solution to mitigate error accumulation during autoregressive inference.

To sum up, our contributions include:
\begin{itemize}
    \item \textbf{Next-detail prediction paradigm}. As shown in \ref{fig:ar:next}, we introduce a novel coarse-to-fine 1D autoregressive image generation framework that progressively refines images from global structures to fine details. This 1D coarse-to-fine token sequence is more aligned with the inference paradigm of autoregressive models.
    \item \textbf{Improved token efficiency}. DetailFlow significantly reduces token requirements, achieving 2.96 gFID using a 326M parameter AR model with only 128 tokens and nearly 2× faster inference speed on the ImageNet 256\(\times\)256 benchmark, compared to 680 tokens required by recent state-of-the-art methods such as VAR \cite{tian2024visual} and FlexVAR \cite{jiao2025flexvar}.
    \item \textbf{Accelerated parallel inference}. Our parallel decoding mechanism combined with a self-correction training strategy boosts inference speed by \(\sim\)8×, simultaneously mitigating the error accumulation typically observed in autoregressive models.
    \item \textbf{Dynamic-resolution 1D tokenization}. DetailFlow uniquely supports dynamic resolution in 1D tokenizers, employing a single 1D tokenizer capable of generating variable-length token sequences, thereby enabling flexible image decoding at multiple resolutions without additional retraining.

  \end{itemize}

\section{Related Work}
\subsection{Image tokenizer} 
The image tokenizer is crucial for generative tasks, as it affects both the quality of generation and the model architecture. An image tokenizer typically encodes high-dimensional visual data into a compressed latent space, representing the image as a sequence of discrete tokens. This process is then reversed by a decoder to reconstruct the original image. Numerous studies have explored and validated this general framework, demonstrating its effectiveness in various model designs.

\textbf{2D image tokenizer.}
Early work such as VQ-VAE \cite{van2017neural, razavi2019generating} introduces a discrete latent representation using vector quantization, enabling token-based image modeling by structuring images as grids of 2D tokens in the latent space. VQGAN \cite{esser2020taming} improves this approach by incorporating adversarial training to improve perceptual quality. Efficient-VQGAN \cite{cao2023efficient} focuses on reducing computational overhead while maintaining reconstruction quality. RQ-VAE \cite{lee2022autoregressive} adopts residual quantization to enrich the representation capacity. 
MoVQ \cite{zheng2022movq} employs a multi-codebook design for more flexible token utilization.

\textbf{1D image tokenizer.}
Unlike 2D tokenizers, which retain spatial structure information in the generated token sequence, the 1D tokenizer TiTok\cite{yu2024image} integrates 2D image information into a 1D token sequence through a self-attention mechanism. This process removes spatial redundancy, thereby enhancing information compression capabilities. However, these tokens lack an inherent order, which poses challenges for their application in next-token prediction within autoregressive models. FlexTok \cite{bachmann2025flextok} addresses this problem by using a tail-drop tokenizer training strategy, which forces information to focus on earlier tokens, thus generating a coarse-to-fine ordered token sequence. However, under a 1.33B AR model, the generation performance, as measured by gFID, deteriorates from approximately 1.9 at 32 tokens to 2.5 at 256 tokens. This limitation, where high-quality generation is only achievable with a small number of tokens, restricts its ability to scale to higher-resolution images that require more tokens for reconstruction. 

\subsection{Visual generation}

\textbf{Autoregressive models.} The definition of an appropriate token sequence is crucial for autoregressive models in image generation tasks. Inspired by NLP, the \textit{next-token prediction} paradigm for sequentially generating discrete image tokens has been naturally extended to decoder-only Transformer architectures. VQGAN \cite{esser2020taming}, Parti \cite{yu2022scaling} and EMU3 \cite{wang2024emu3} adopt a raster-scan strategy to arrange 2D image tokens into a 1D ordered token sequence by scanning row by row. 
While this method facilitates token-level prediction in a manner analogous to text generation, alternative approaches have explored more structured generation schemes. Notably, VAR \cite{tian2024visual} introduces a \textit{next-scale prediction} framework, which departs from the conventional sequential token prediction by generating coarser-to-finer representations across multiple scales. This hierarchical strategy enables the model to capture global image structure before refining local details. Similarly, CART \cite{roheda2024cart} proposes a "next-detail" prediction strategy that compositionally generates an image by first predicting a "base" factor, which captures the global structure, and then iteratively adding "detail" factors to refine local features.

\textbf{Masked-prediction model.} MaskGIT \cite{chang2022maskgit} is a transformer-based image synthesis framework that leverages masked visual token modeling to generate images through parallel decoding. Unlike autoregressive models, it predicts all tokens simultaneously and iteratively refines them. TiTok \cite{yu2024image} uses it for image generation with 1D token sequence. MUSE \cite{chang2023muse} extends MaskGIT to text-to-image generation by integrating a pretrained language model with MaskGIT's parallel decoding.

\textbf{Diffusion models} synthesize data through a forward-backward stochastic process: gradually corrupting data with Gaussian noise and learning to reverse this degradation via iterative denoising. Latent diffusion models \cite{rombach2022high} first proposes modeling in the latent space using a U-Net architecture. DiT \cite{peebles2023scalable} replaces conventional U-Net with a transformer architecture for latent image processing, demonstrating superior scalability and achieving better image generation quality. 

\section{Method}

\begin{figure}
    \begin{center}
        \includegraphics[width=0.99\linewidth]{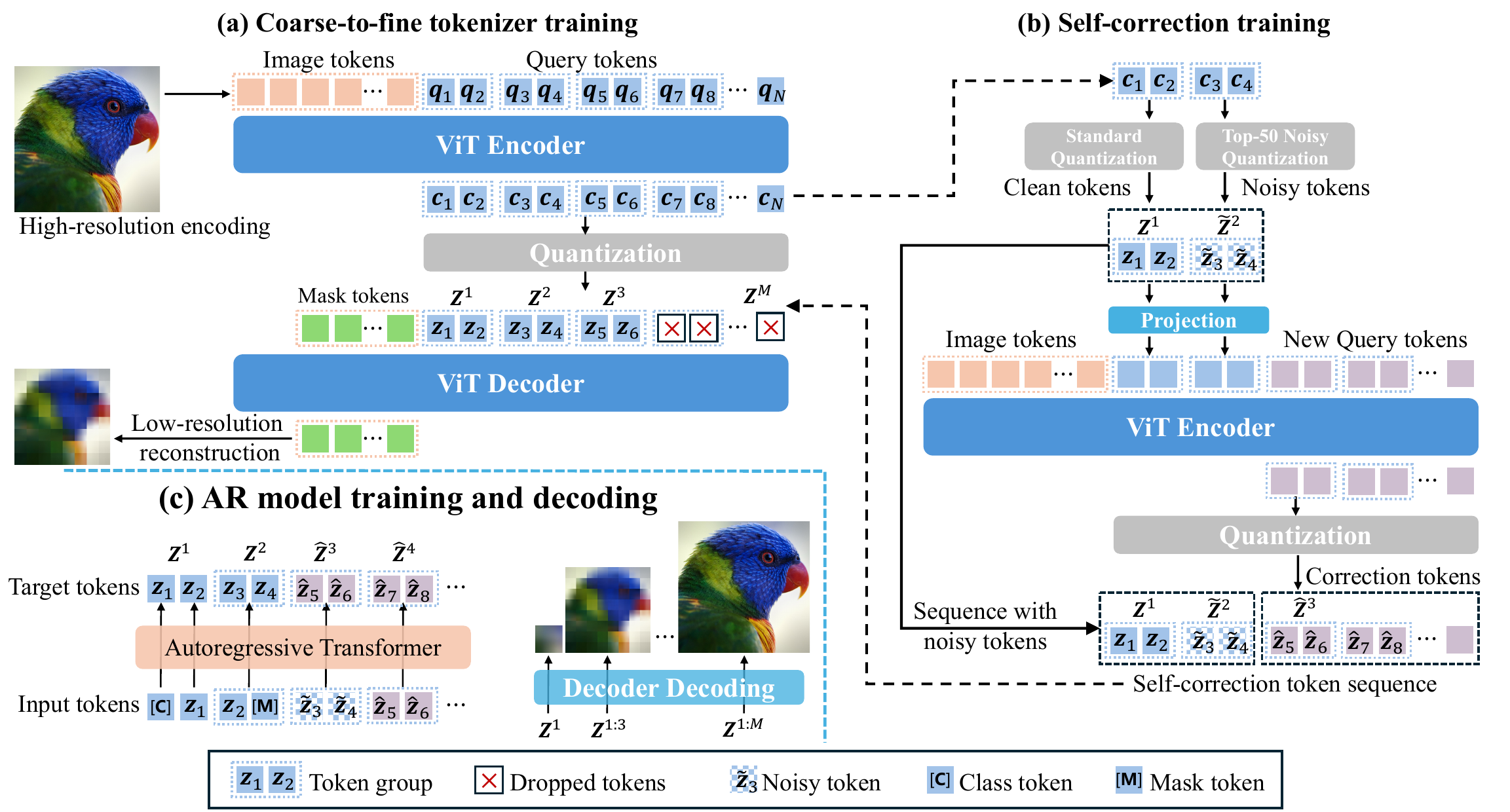}
     \end{center}
    \caption{(a) Coarse-to-fine tokenizer training. The encoder maps high-resolution images to 1D latent token sequences. Decoding with more tokens yields higher-resolution outputs, with earlier tokens capturing global structure and later ones refining details. (b) Self-correction training. Randomly perturbed tokens are re-encoded, and encourages subsequent tokens to correct errors from earlier noisy tokens. (c) Autoregressive (AR) model training and decoding. AR model predicts the first group of tokens in a next-token prediction manner, followed by parallel prediction of subsequent groups. At inference, more predicted tokens lead to higher-resolution outputs.}
    \label{fig:arc:overview}
  \end{figure}

Encoding images into a 2D grid of tokens is a natural and widely adopted strategy, leveraging the inherent spatial structure of images. However, using a 1D tokenizer allows for more efficient compression, representing images with fewer tokens and offering greater flexibility in controlling the information content of each token. This flexibility is critical for balancing image quality and computational efficiency.

\subsection{Preliminary Background on 1D Tokenizer}
The general structure of tokenizers typically consist of an encoder, a quantizer, and a decoder. The encoder embeds an input image into continuous image tokens, the quantizer discretizes them into discrete tokens, and the decoder reverses this process to reconstruct the original image. Similarly, the 1D tokenizer adheres to this paradigm but introduces slight modifications to the encoder and decoder.

Given an input image \(\mathbf{X} \in \mathbb{R}^{H \times W \times 3}\), it is first patchified to non-overlapping patches \(\mathbf{P} \in \mathbb{R}^{\frac{H}{f} \times \frac{W}{f} \times D}\), where \(f\) is the patch size and \(D\) is the patch feature dimension. These patches are concatenated with a set of learnable 1D query tokens \(\mathbf{Q} \in \mathbb{R}^{N \times D}\) and fed into the encoder. \(N\) is the number of query tokens. The encoder outputs continuous latent tokens \(\mathbf{C} \in \mathbb{R}^{N \times D}\), which are then discretized by a quantizer into discrete tokens \(\mathbf{Z} \in \mathbb{R}^{N \times d}\) using a codebook of dimension \(d\).

For reconstruction, the discrete tokens \(\mathbf{Z}\) are first projected and then concatenated with learnable mask tokens \(\mathbf{M} \in \mathbb{R}^{\frac{H}{f} \times \frac{W}{f} \times D}\) \cite{baobeit}, created by duplicating a single mask embedding \(\mathbf{m} \in \mathbb{R}^{1 \times D}\). The decoder output corresponding to the mask tokens is regressed to the pixel values through a linear projection. 2D position embeddings are applied for image tokens and mask tokens. 1D position embeddings are applied for query tokens and latent tokens.  

\subsection{Coarse-to-Fine 1D Latent Representation}

Human perception and image creation are inherently hierarchical processes, starting with a rough global structure and progressively refining local details. This hierarchical approach reduces complexity and improves quality by breaking down the task into simpler steps. Inspired by this, we design a coarse-to-fine information ordering for 1D latent tokens, enabling the model to progressively generate images from global to fine-grained details.

As illustrated in Fig.~\ref{fig:arc:overview}(a), we enforce an information ordering in the 1D latent space by leveraging the correlation between image resolution and semantic granularity: lower-resolution images primarily preserve global structure, while higher-resolution images capture increasingly detailed content. We define a resolution mapping function \(\mathcal{R}(n)\) linking the number of used tokens \(n\) to a target resolution \(r_n = \sqrt{hw} = \mathcal{R}(n)\). Early tokens are trained to capture coarse structures at low resolutions, while later tokens refine high-frequency details. To enforce this, we use causal (unidirectional) attention among query tokens in the encoder, while maintaining bidirectional attention for image tokens in the encoder and all the tokens in decoder.

During training, we randomly sample \(n \in [1, N]\) and reconstruct a downsampled version of \(\mathbf{X}\) at resolution \(\mathcal{R}(n)\) using only the first \(n\) latent tokens \(\mathbf{Z}_{1:n} = \{\mathbf{z}_1, \mathbf{z}_2, \dots, \mathbf{z}_n\}\). Mask tokens and positional embeddings of the decoder are adjusted according to the target image size. The model is supervised to reconstruct downsampled image from this partially observed latent sequence, ensuring that the earlier tokens specialize in capturing global structure while later tokens incrementally contribute finer details.

Formally, the conditional entropy of the $i$-th token, \(H(\mathbf{z}_i \mid \mathbf{Z}_{1:i-1})\), quantifies the incremental information it contributes. The total entropy up to token \(n\) is:
\begin{equation}
H(r_n) = \sum_{i=1}^n H(\mathbf{z}_i \mid \mathbf{Z}_{1:i-1}).
\end{equation}
By leveraging this hierarchical decomposition, our method ensures that each token contributes meaningfully to the reconstruction of image details, enabling a progressive and efficient representation of high-dimensional image data.

Assuming the entropy per pixel is $H(r=1)$, the total image entropy at resolution $r \times r$ scales as:
\begin{equation}
\label{eq:info:h}
H(r) \propto r^2 H(r=1),
\end{equation}
indicating a nonlinear relationship between the number of tokens and reconstructable resolution. To model this, we define \(\mathcal{R}(n)\) (shown in  Fig.~\ref{fig:abla:fid}(b)) as:
\begin{align}
    \mathcal{R}(n) = R - b(N-n)^\alpha = R - \underbrace{\frac{R - 1}{(N - 1)^\alpha}}_{b}(N - n)^\alpha,\ \text{where}\ \mathcal{R}(1) = 1,\ \mathcal{R}(N) = R.
\end{align}
\(R\) is the maximum resolution supported by the tokenizer, and coefficient \(b\) is calculated using the condition \(\mathcal{R}(1) = 1\), ensuring the first token corresponds to a resolution of \(1 \times 1\). The condition \(\mathcal{R}(N) = R\) ensures the total \(N\) tokens corresponds to the maximum resolution \(R\). The hyperparameter \(\alpha\) controls the degree of nonlinearity. The continuous resolution value \( \mathcal{R}(n)\) is rounded to the nearest multiple of \(f\) to obtain the final target resolution for compatibility with the decoder's patch size \(f\). For simplicity, this step is omitted in formulas and visualizations.

\subsection{Parallel Inference Acceleration}

Generating high-resolution images often requires thousands of tokens, making purely sequential next-token prediction inefficient. To address this, we partition the 1D token sequence into \(M\) groups of \(g\) tokens each. During tokenization, we apply bidirectional attention within groups and causal attention across groups. For coarse-to-fine training, we randomly sample an integer \(k \in [1, M] \) and reconstruct the downsampled image using the first \(k\) groups \(\mathbf{Z}^{1:k} = \{ \mathbf{Z}^1, \mathbf{Z}^2, \ldots, \mathbf{Z}^k \}\).

For the autoregressive (AR) model, considering that the first group controls the global structure, which is extremely important, we keep the tokens in the first group as causal attention \cite{wang2024parallelized} and use next-token prediction as shown in Fig.~\ref{fig:arc:overview}(c). Subsequent groups use bidirectional attention internally and causal attention between groups, enabling parallel prediction of \(g\) tokens per step.

However, the independent sampling of tokens within a group during parallel inference disrupts intra-group dependencies, introducing sampling errors. The teacher-forcing training paradigm does not equip the AR model with the ability to self-correct such errors. To mitigate this, prior works \cite{lee2022autoregressive, wu2024vila} introduce a depth transformer head to sequentially predict tokens within a group, but at the cost of model complexity. Alternatively, Infinity \cite{han2024infinity} simulates self-correction by randomly flipping a quantized value for one scale and recalculating the residuals at subsequent scales, enabling the AR model to learn self-correction capabilities from self-correction token sequences.

Inspired by these ideas, we need to obtain a 1D self-correction token sequence where a sampling error occurs at one position, and the subsequent token sequence can correct the sampling error.
Therefore, we inject stochastic perturbations into the quantization process. Specifically, as shown in Fig.~\ref{fig:arc:overview}(b) we randomly select a token group \(\mathbf{C}^{m}, m \in [1, k-1] \) and, during quantization, sample each token from the top-50 nearest codebook entries, yielding a noisy group \(\mathbf{\tilde{Z}}^m\). To obtain the subsequent correlation tokens, we introduce a correction mechanism: feeding the clean tokens \(\mathbf{Z}^{1:m-1} \), noisy tokens \(\mathbf{\tilde{Z}}^m\) (with gradient truncation), and new query tokens back into the encoder. A projection layer ensures dimensional compatibility. The encoder then generates corrected tokens \(\{ \mathbf{\hat{Z}}^{m+1}, \ldots, \mathbf{\hat{Z}}^{M}\}\), forming a self-correction sequence \(\{ \mathbf{Z}^{1:m-1}, \mathbf{\tilde{Z}}^m, \mathbf{\hat{Z}}^{m+1:k} \}\).

To jointly train standard sequence modeling and self-correction, we concatenate the original sequence \(\mathbf{Z}^{1:k}\) and the self-correction sequence \(\{ \mathbf{Z}^{1:m-1}, \mathbf{\tilde{Z}}^m, \mathbf{\hat{Z}}^{m+1:k} \}\) of the same input image along the batch dimension, independently decoding both to reconstruct the same target image. This trains the encoder to learn to output subsequent token sequences that can correct sampling errors in the preceding tokens. This design enables the tokenizer to provide both standard and self-correction sequences simultaneously, which enables the AR model to learn self-correction capabilities alongside standard token modeling, improving image quality during parallel inference.

\subsection{Training Objective}
Since early tokens encode global structure, we explicitly align the first latent token \(\mathbf{z}_1\) with the globally pooled features extracted by the pre-trained Siglip2 model \cite{tschannen2025siglip}. This allows the first token to better embed global information. Specifically, the first token \(\mathbf{z}_1\) is projected through a three-layer MLP and aligned via cosine similarity as the alignment loss:
\begin{equation}
    \mathcal{L}_{\text{align}} = - \cos\left(\text{MLP}(\mathbf{z}_1), \text{Siglip2}(\mathbf{X})\right).
    \end{equation}    
In addition to the alignment loss, the final training objective of the tokenizer also includes the reconstruction loss, perceptual loss \cite{dosovitskiy2016generating, johnson2016perceptual}, adversarial loss \cite{goodfellow2020generative, isola2017image} and VQ codebook loss \cite{esser2021taming}, following the implementations and weighting schemes used in SoftVQ-VAE \cite{chen2024softvq}.

\section{Experiments}
\subsection{Implementation}
\textbf{Tokenizer Setup.} 
The encoder including 12 layers is initialized with the weigths of Siglip2-NaFlex \cite{tschannen2025siglip}, yielding a parameter count of 184M. In contrast, the decoder is trained from scratch, comprising 86M parameters. The discrete latent space is defined by a codebook with 8,192 entries and a dimension of 8. 
Tokenizer training is conducted on the ImageNet-1K \cite{deng2009imagenet}, using \(256 \times 256\) resolution inputs to the encoder and dynamically varying output resolutions (up to 
\(256 \times 256\)) from the decoder. 
We train the tokenizer for 250 epochs with a batch size of 256 using a cosine learning rate decay strategy, starting from an initial learning rate of 1e-4. To ensure robust modeling of the entire latent token sequence, we reconstruct full-resolution images (\(n=N\)) with an 80\% probability, and with a 20\% probability, we randomly reconstruct lower-resolution images by sampling the first \(n=kg\) tokens, where \(k \in [1, M-1], g=8, M=N/g\). We achieve full codebook utilization (100\%) across all tokenizers.

\textbf{AR Model Setup.}
For downstream generation, we adopt an autoregressive (AR) model based on the LlamaGen architecture \cite{sun2024autoregressive}. 
The AR model is trained on ImageNet-1K using a cosine-decayed learning rate schedule for 300 epochs. 30\% of the training data consists of curated self-correction token sequences. During inference, we apply sampling with Top-K=8192 and Top-P=1. Classifier-Free Guidance (CFG) is tuned to its optimal value to balance generation diversity and fidelity. The default CFG value is 1.5. The inference time is measured on a single A100 using the batch size 1.

\textbf{Metrics.} We evaluate image reconstruction quality on the ImageNet-1K validation set using rFID \cite{heusel2017gans}, Peak Signal-to-Noise Ratio (PSNR), and Structural Similarity Index (SSIM). For assessing image generation quality, we report Frechet Inception Distance (FID) \cite{heusel2017gans}, Inception Score (IS) \cite{salimans2016improved}, as well as Precision and Recall metrics. 

\subsection{State-of-the-art image generation}

\begin{table}
    \caption{Comparison of class-conditional image generation on ImageNet-1k at \(256 \times 256\) resolution.
    Models marked with \(^\dag\) leverage additional training data beyond ImageNet. The \(^\ddagger\) symbol indicates methods that do not employ classifier-free guidance. The \(^\diamond\) symbol denotes models that generate images at \(384 \times 384\) resolution, which are subsequently downsampled to \(256 \times 256\) for evaluation.
    \textbf{\#Tokens} represents the total number of tokens predicted during autoregressive (AR) model training. \textbf{Flex} indicates whether the AR model supports dynamic resolution, meaning it can decode images of different resolutions using varying numbers of tokens. For DetailFlow, the notation 16*8 indicates that the 128 tokens are partitioned into 
 \(M = 16\) groups, each containing \(g=8\) tokens. Reported evaluation metrics include rFID and gFID. The inference steps of FlexTok \cite{bachmann2025flextok} include the steps of the AR model and the Diffusion model.}
    \label{tab:sota:all}
    \centering
    \setlength{\tabcolsep}{0.3mm}{
    \begin{tabular}{clc|lccccccc}
        \hline
        Type & Tokenizer & rFID\(\downarrow\) & Generator & Type & Param. & \#Tokens & gFID\(\downarrow\) & Step & Time(s) & Flex \\
        \hline
        \multicolumn{10}{c}{Continuous modeling} \\
        \hline
        2D & GigaGAN\cite{kang2023scaling} & - & GigaGAN\cite{kang2023scaling} & GAN & 569M & - & 3.45 & \textbf{1} & - & \ding{51}  \\
        2D & VAE\(^\dag\)\cite{rombach2022high} & \textbf{0.27} & LDM-4\cite{rombach2022high} & Diff. & 400M & 4096 & 3.60   & 250 & -  & \ding{51} \\
        2D & SD-VAE\cite{sdvae} & 0.62 & SiT-XL/2\cite{ma2024sit} & Diff. & 675M & 1024  & 2.06  & 250 & - & \ding{51} \\
        2D & VAE\cite{li2024autoregressive} & 0.53 & MAR-H\cite{li2024autoregressive} & AR+Diff. & 943M & 256 & \textbf{1.55} & 64 & 28.24 & \ding{51} \\
        \hline
        \multicolumn{10}{c}{Discrete modeling} \\
        \hline
        1D & TiTok-S\cite{yu2024image} & 1.71 & MaskGIT\cite{chang2022maskgit} & Mask. & 287M & 128  & 1.97  & 64 & 0.13  & \ding{55} \\
        \multirow{2}{*}{1D} & \multirow{2}{*}{FlexTok\cite{bachmann2025flextok}} & \multirow{2}{*}{1.45} & \multirow{2}{*}{LlamaGen\cite{sun2024autoregressive}} & \multirow{2}{*}{AR+Diff.} & \multirow{2}{*}{1.33B} & 32  & 1.86\(^\ddagger\) & 57 & - & \ding{55} \\
         &  &  & &  &  & 256  & \(\sim\!2.5^\ddagger\) & 281 & - & \ding{55} \\
        2D & ImageFolder\cite{liimagefolder} & 0.80 & VAR-GPT\cite{tian2024visual} & VAR  & 362M & 286*2  & 2.60 & 10 & 0.13 & \ding{55} \\
        \hline
        2D & VQGAN\(^\dag\)\cite{chang2022maskgit} & 2.28 & MaskGIT\cite{chang2022maskgit} & Mask. & 227M & 256 & 6.18\(^\ddagger\)  & 8 & 0.13  & \ding{51} \\
        2D & ViT-VQGAN\(^\dag\)\cite{yu2021vector} & 1.28 & VIM-Large\cite{yu2021vector} & AR & 1.7B & 1024 & 4.17\(^\ddagger\) & 1024 & >6.38  & \ding{51} \\
        2D & RQ-VAE\cite{lee2022autoregressive} & 3.20 & RQTran.-re\cite{lee2022autoregressive} & AR & 3.8B & 256 & 3.8\(^\ddagger\) & 64 & 5.58 & \ding{51} \\
        2D & LlamaGen\cite{sun2024autoregressive} & 2.19 & GPT-L\cite{sun2024autoregressive} & AR & 343M & 256 & 3.8 & 256 & 12.58  & \ding{51} \\
        2D & O.-MAGVIT2\cite{luo2024open} & 1.17 & AR-B\cite{luo2024open} & AR & 343M & 256  & 3.08 & 256 & - & \ding{51} \\
        2D & PAR\cite{wang2024parallelized} & 0.94 & PAR-L-4\cite{wang2024parallelized} & AR  & 343M & 576 & 3.76\(^\diamond\)  & 147 & 3.38 & \ding{51} \\
        2D & VAR\(^\dag\)\cite{tian2024visual} & 0.90 & VAR-d16\cite{tian2024visual} & VAR  & 310M & 680 & 3.30  & 10 & 0.15 & \ding{51} \\
        2D & FlexVAR\(^\dag\)\cite{jiao2025flexvar} & - & VAR-d16\cite{jiao2025flexvar} & VAR & 310M & 680 & 3.05  & 10 & 0.15 & \ding{51} \\
        
        \hline
        \rowcolor{gray!10}
        1D & DetailFlow-16 & 1.22 & GPT-L \cite{sun2024autoregressive} & AR & 326M & 16*8  & 2.96 & 23 & \textbf{0.08} & \ding{51} \\
        \rowcolor{gray!10}
        1D & DetailFlow-32 & 0.80 & GPT-L \cite{sun2024autoregressive} & AR & 326M & 32*8  & 2.75 & 39 & 0.16  & \ding{51} \\
        \rowcolor{gray!10}
        1D & DetailFlow-64 & 0.55 & GPT-L \cite{sun2024autoregressive} & AR & 326M & 64*8  & 2.62 & 71 & 0.38 & \ding{51} \\
        \hline
      \end{tabular}}
  \end{table}

  \begin{figure}[t]
    \begin{center}
        \includegraphics[width=0.30\linewidth]{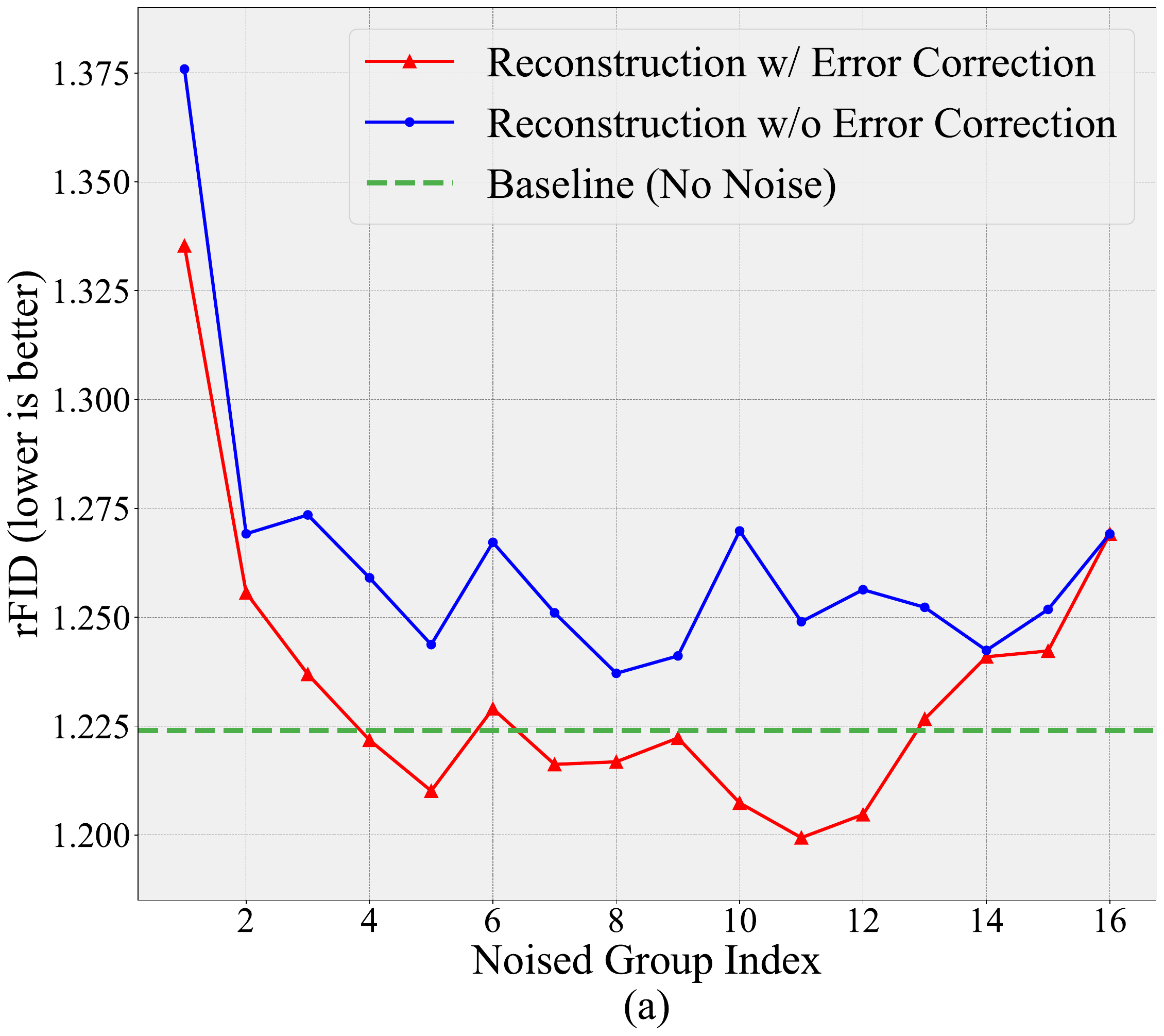}
        \hspace{0.15cm}
        \includegraphics[width=0.32\linewidth]{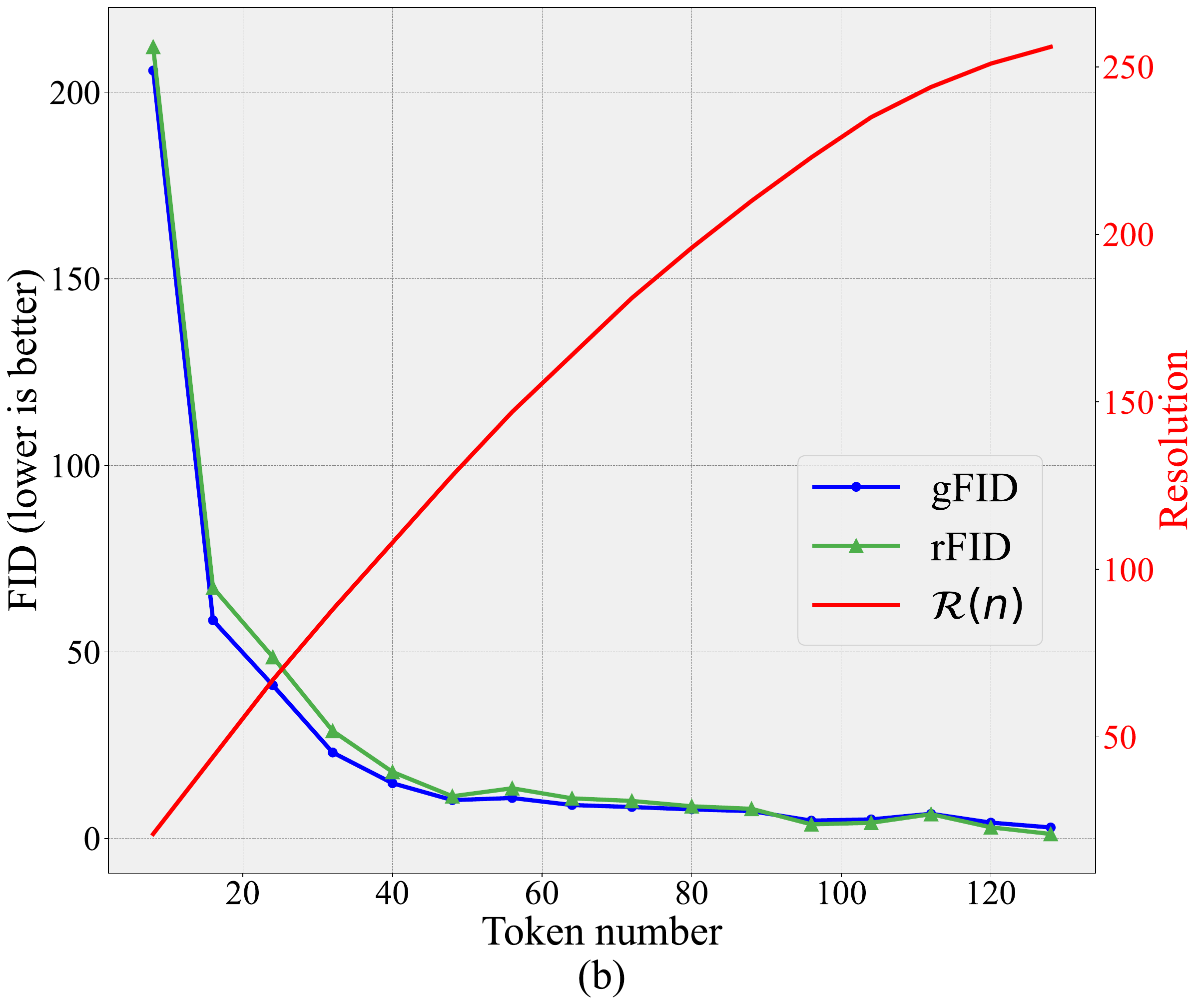}
        \hspace{0.15cm}
        \includegraphics[width=0.29\linewidth]{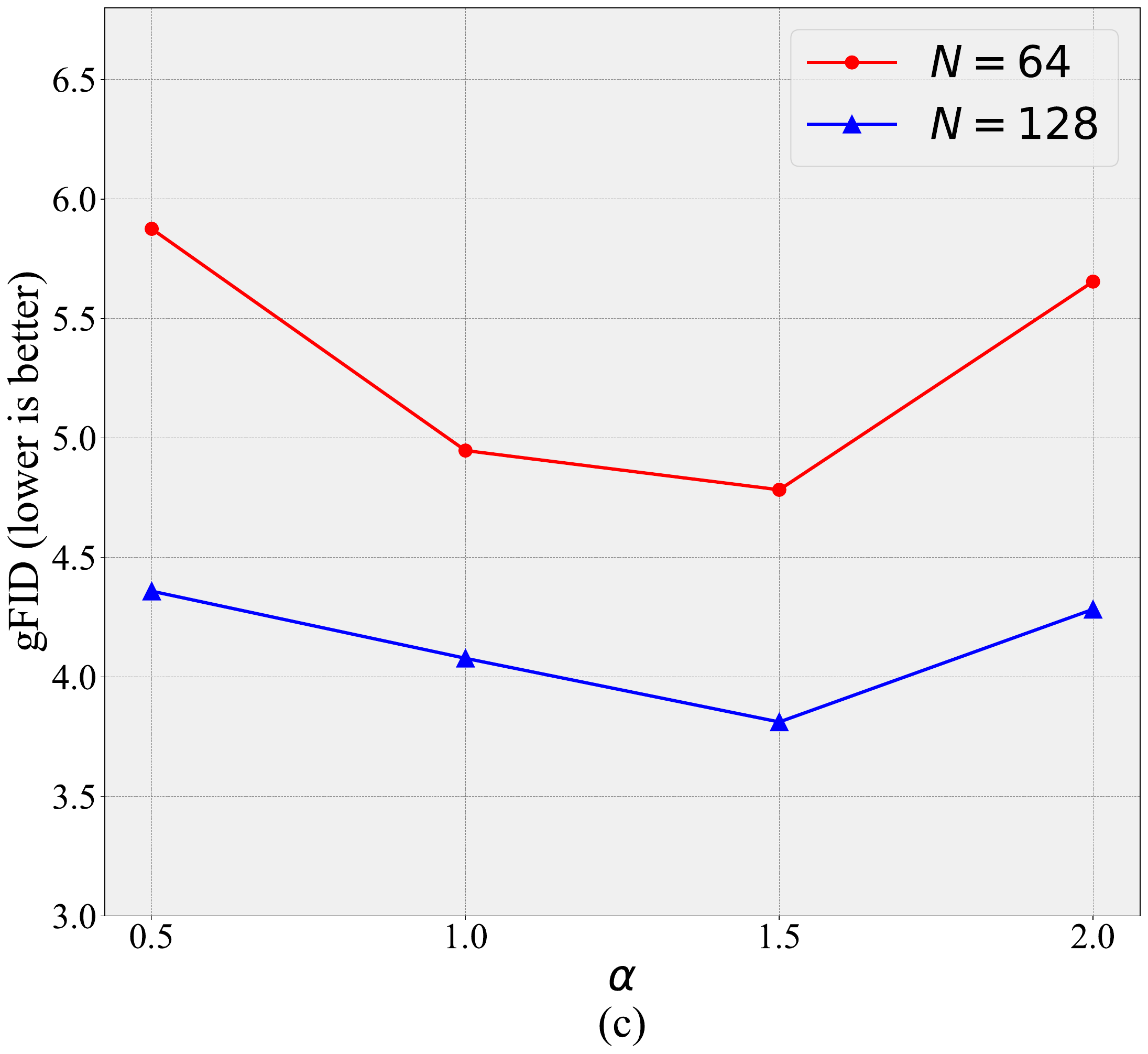}
        \end{center}
    \caption{(a) Reconstruction metrics before and after self-correction when adding noise to latent tokens of a group (tokenizer with 128 tokens, group size 8, trained for 200 epochs). (b)
    Impact of token count on image resolution, reconstruction quality (rFID), and generation quality (gFID), with all evaluations conducted on images resized to \(256 \times 256\). The tokenizer is identical to (a). (c) Influence of the hyperparameter \(\alpha\) in the mapping function \(\mathcal{R}(n)\) on generation metrics, using tokenizers trained for 50 epochs.}
    \label{fig:abla:fid}
    \end{figure}

In Table~\ref{tab:sota:all}, we evaluate our proposed method, DetailFlow, on the ImageNet \(256 \times 256\) benchmark, comparing it against a range of state-of-the-art generative models—including GANs \cite{kang2023scaling}, diffusion models \cite{rombach2022high, sdvae, li2024autoregressive}, masked prediction models \cite{chang2022maskgit, bachmann2025flextok}, and autoregressive (AR) models \cite{sun2024autoregressive, luo2024open, tian2024visual, jiao2025flexvar}. 

  \textbf{Compared to existing 2D tokenizers} \cite{sun2024autoregressive, luo2024open, tian2024visual, jiao2025flexvar} that support dynamic resolution, our method delivers higher quality with shorter sequence length among ar models. DetailFlow-16 achieves a lower gFID of 2.96 using only 128 tokens, surpassing VAR (3.3 FID) \cite{tian2024visual} and FlexVAR (3.05 FID) \cite{jiao2025flexvar}, which both require 680 tokens. Additionally, its reduced token count and parallel inference make it nearly twice as fast as VAR and FlexVAR during inference. This is attributed to the fact that the 1D tokenizer effectively eliminates spatial information redundancy, allowing more information to be carried with fewer tokens. Although ImageFolder \cite{liimagefolder} reports a gFID score comparable to ours, its lack of support for dynamic resolution constrains its practical applicability. While ImageFolder employs a 1D tokenizer architecture, the resulting latent tokens are still arranged in a 2D structure and preserve explicit spatial information, limiting it to decoding images at a fixed resolution.

  \textbf{Compared to existing 1D tokenizers} \cite{yu2024image, bachmann2025flextok}, our approach addresses several key limitations. Notably, prior 1D tokenizers do not support multi-resolution image generation and parallel inference with self-correction mechanisms. Additionally, TiTok \cite{yu2024image} lacks an explicit and structured token ordering necessary for effective AR modeling, while FlexTok \cite{bachmann2025flextok} demonstrates limited scalability, with performance degrading as token count increases. For instance, using a 1.33B AR model, its gFID score rises from roughly 1.9 at 32 tokens to about 2.5 at 256 tokens.  In contrast, DetailFlow supports coarse-to-fine image generation (Fig.~\ref{fig:demo:generation}(a)), allowing for the prediction of more tokens to decode higher-resolution images. It also supports parallel inference with self-correction to accelerate the image generation process, making it both scalable and efficient.

\subsection{Ablation Study}

  \begin{wraptable}{r}{0.53\textwidth}
    \caption{Ablation study of DetailFlow. These tokenizers with \(N=128\) latent tokens are trained with only 50 epochs. Reported evaluation metrics include rFID, PSNR, gFID and Recall.}
    \label{tab:abla:arc}
    \centering
    \setlength{\tabcolsep}{0.3mm}{
    \begin{tabular}{l|cc|ccc}
        \hline
        Setting & rFID\(\downarrow\) & PSNR\(\uparrow\) & gFID\(\downarrow\) & Rec\(\uparrow\) & Step \\
        \hline
        Baseline & 1.73 & 19.47 & 3.97 & 0.50 & 128 \\
        +causal encoder & 1.87 & 19.49 & 3.66 & 0.53 & 128 \\
        +coarse-to-fine & 1.92 & 19.31 & 3.33 & 0.54 & 128 \\
        +parallel\((g=4)\) & 1.87 & 19.27 & 4.11 & 0.50 & 32 \\
        +self-correction & \multirow{2}{*}{1.81} & \multirow{2}{*}{19.16} & 3.68 & 0.51  & 32 \\
        +first group causal &  &  & 3.59 & 0.51  & 35 \\
        +alignment loss & 1.68 & 19.05 & 3.35 & 0.55 & 35 \\
        \hline
      \end{tabular}}
\end{wraptable}

 In Fig.~\ref{fig:abla:fid}(a), we analyzes the reconstruction performance before and after self-correction when noise is injected into different groups of latent tokens. When noise is applied to earlier token groups (\textit{i.e.}, smaller group indices), reconstruction quality deteriorates sharply. This is expected, as early tokens encode the global structure, and inaccuracies in this region propagate large-scale distortions throughout the image. Moreover, in these cases, the self-correction mechanism exhibits limited effectiveness, since later tokens—responsible mainly for fine details—lack the capacity to rectify global errors. As the group index increases, the model demonstrates strong ability to correct errors introduced by noisy tokens. However, when noise is injected into the final few groups, the correction performance again deteriorates, as fewer subsequent tokens are available to facilitate error correction.

These results underscore the importance of early tokens for capturing global structure. To enhance their reliability, our model incorporates self-correction training, causal modeling for the first token group, and an alignment loss on the initial token.
We further perform an ablation study to evaluate each component‘s contribution. Starting from a baseline that encodes images into an unordered token sequence, we progressively add modules to measure their effects.

First, introducing a causal encoder establishes a simple sequential order among the tokens, which substantially improves the model’s capacity for autoregressive generation. Building on this, we implement a coarse-to-fine tokenizer training strategy by supervising reconstructions at multiple resolutions. The observed improvements in the gFID metric from 3.66 to 3.33 validate that enforcing such coarse-to-fine semantic ordering is both effective and advantageous.

\begin{wrapfigure}{r}{0.5\textwidth} 
    \includegraphics[width=0.49\textwidth]{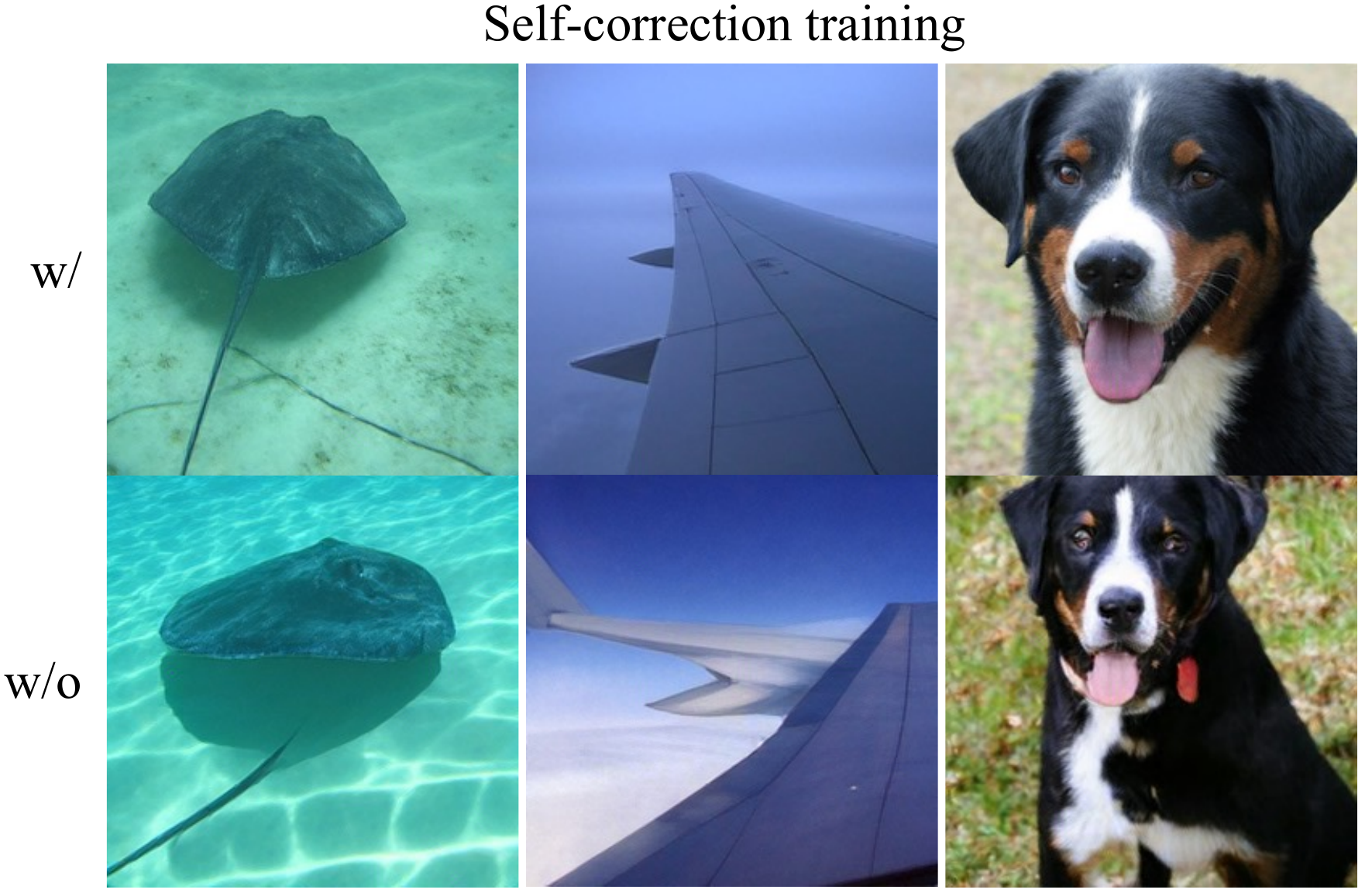} 
    \caption{Qualitative comparison of AR model outputs with (w/) and without (w/o) self-correction training.}
    \label{fig:demo:case}
\end{wrapfigure}

Next, we explore parallel prediction of token groups. Although this design reduces the inference steps from 128 to 32, it introduces degradation in generation quality, primarily due to the accumulation of sampling errors across groups. To mitigate this, we incorporate a self-correction mechanism, which substantially restores synthesis quality by allowing the model to iteratively refine predictions. This reduces the gFID score from 4.11 to 3.68. Fig.~\ref{fig:demo:case} shows that self-correction training improves global structure and detail quality, suggesting that self-correction effectively mitigates the impact of sampling errors.

Further enhancement is achieved by applying causal next-token prediction specifically to the first group of tokens, yielding an additional improvement of 0.09 in gFID. This stabilizes the generation process and improves the fidelity of the final output. 
Finally, to strengthen global semantic consistency, we align the first token's representation with Siglip2 global image features via an alignment loss. This further lowers the gFID from 3.59 to 3.35, indicating that anchoring the initial token to global structural information provides stronger guidance for the entire generation process.

Fig.~\ref{fig:abla:fid}(b) shows how increasing the number of latent tokens impacts output resolution, reconstruction metrics, and generation metrics. As the number of tokens grows, the latent sequence encodes finer-grained information, facilitating higher-resolution image decoding. Both reconstruction and generation metrics improve correspondingly, with the autoregressive model benefiting from richer detail prediction, leading to enhanced image quality and resolution.

Fig.~\ref{fig:abla:fid}(c) examines the effect of hyperparameter \(\alpha\) in the mapping function \(\mathcal{R}(n)\) on generation metrics. The parameter \(\alpha\) governs the relationship between token quantity and decoding resolution. Smaller values of \(\alpha\) are preferable, as they indicate that fewer additional tokens are required to support higher-resolution images. Across different total token counts, the optimal value of \(\alpha\) is consistently 1.5. This value, greater than 1, reflects that higher-resolution regions demand more tokens compared to lower-resolution regions, aligning with Eq.~\ref{eq:info:h}. Meanwhile, \(\alpha\) being less than 2 highlights the tokenizer's ability to compress spatial redundancy inherent in the image data.

\section{Conclusion}

In this paper, we propose a novel  autoregressive image generation method DetailFlow, introducing a new image generation paradigm called Next-Detail Prediction. By leveraging a 1D tokenizer trained on progressively degraded images, DetailFlow establishes a direct correspondence between token sequences and image resolution levels, enabling a coarse-to-fine generation strategy that enhances visual fidelity. This method effectively compresses image information into a smaller token sequence while maintaining high-quality image generation. Furthermore, a parallel decoding mechanism with self-correction improves inference speed without compromising image quality. Overall, DetailFlow achieves an effective balance among training cost, inference efficiency, and image quality, offering a scalable solution for high-resolution, autoregressive image synthesis.

\bibliographystyle{plain}
\bibliography{neurips_2025}


\appendix

\section{Technical Appendices and Supplementary Material}

\subsection{Implementation Details}

\subsubsection{Tokenizer Setup}

We initialize the encoder using the SigLIP2-NaFlex weights and retain only the first 12 layers to reduce memory consumption, resulting in an encoder of 184M parameters. As shown in Fig.~\ref{fig:arc:attention}(a), the attention design of the encoder includes bidirectional attention within image tokens and within each latent token group (except the first), causal attention within the first latent group, and causal attention across groups. 

\subsubsection{AR model Setup}

\begin{wrapfigure}{r}{0.6\textwidth} 
    \includegraphics[width=0.59\textwidth]{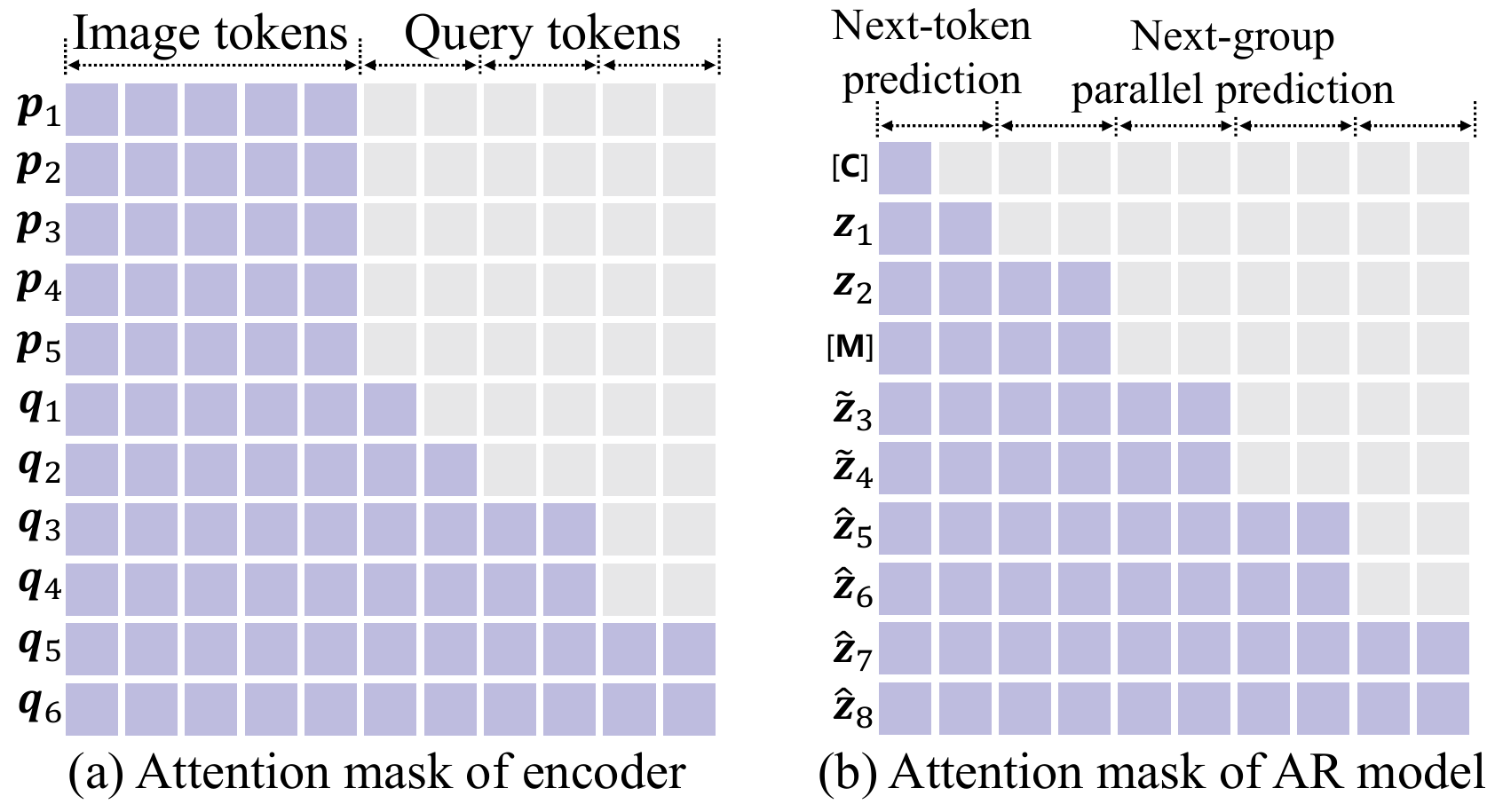} 
    \caption{The attention mask in the proposed method. We use a group size \(g=2\) for latent tokens as an example.}
    \label{fig:arc:attention}
\end{wrapfigure}

Building on the autoregressive LlamaGen architecture, we incorporate learnable mask tokens inspired by PAR \cite{wang2024parallelized} to enable parallel decoding. We also use the 2D positional encoding to enhance the model's ability to distinguish between different token groups and intra-group tokens. As illustrated in Fig.~\ref{fig:arc:attention}(b), the attention pattern of the AR model is structured such that the first group uses causal attention for next-token prediction, while subsequent groups apply bidirectional attention internally and causal attention across groups. This configuration allows for parallel prediction of later groups while preserving autoregressive dependencies. Training is conducted primarily on 16 A800 GPUs.

\subsection{Additional Results}

\subsubsection{Additional Ablation Study}

\textbf{Impact of Coarse-to-fine Training Probability.} During the training process of DetailFlow, we introduce a probabilistic strategy to balance between full-resolution and degraded-resolution image reconstruction. Specifically, with a certain probability, the model is trained with complete token sequences to reconstruct full-resolution images, while with the remaining probability, it is trained on downsampled images with partial tokens to encourage learning the coarse-to-fine ordering of token representations.
 \begin{table}[t]
    \caption{The impact of coarse-to-fine training probability on reconstruction and generation metrics. These tokenizers with \(N=128\) latent tokens are trained with only 50 epochs and the GPT-L AR model are trained with 100 epochs. Reported evaluation metrics include rFID, PSNR, SSIM, gFID, sFID, Inception Score(IS), Precision (Pre) and Recall (Rec).}
    \label{tab:abla:prob}
    \centering
    \setlength{\tabcolsep}{1.7mm}{
    \begin{tabular}{c|ccc|ccccc}
        \hline
        Probability & rFID\(\downarrow\) & PSNR\(\uparrow\) & SSIM\(\uparrow\) & gFID\(\downarrow\) & sFID\(\downarrow\) & IS\(\uparrow\) & Pre\(\uparrow\) & Rec\(\uparrow\) \\
        \hline
        10\% & 2.0 & 19.3 & 0.61 & 4.18 & 5.60 & 202.6 & 0.85 & 0.51 \\
        20\% & 1.8 & 19.2 & 0.60 & 3.81 & 5.46 & 230.1 & 0.86 & 0.51 \\
        30\% & 1.9 & 19.3 & 0.61 & 4.18 & 5.90 & 215.3 & 0.86 & 0.49 \\
        \hline
      \end{tabular}}
      \vspace{1em}
    \caption{The impact of group size \(g\) on reconstruction and generation metrics. These tokenizers with \(N=256\) latent tokens are trained with 250 epochs and the GPT-L AR model are trained with 300 epochs. Reported evaluation metrics include rFID, PSNR, SSIM, gFID, sFID, Inception Score(IS), Precision (Pre) and Recall (Rec).}
    \label{tab:abla:g}
    \centering
    \setlength{\tabcolsep}{1.7mm}{
    \begin{tabular}{l|ccc|cccccc}
        \hline
        Group & rFID\(\downarrow\) & PSNR\(\uparrow\) & SSIM\(\uparrow\) & gFID\(\downarrow\) & sFID\(\downarrow\) & IS\(\uparrow\) & Pre\(\uparrow\) & Rec\(\uparrow\) & Step \\
        \hline
        32*8 (\(g=8\)) & 0.80 & 20.8 & 0.67 & 2.75 & 5.77 & 250.8 & 0.81 & 0.58 & 39 \\
        16*16 (\(g=16\)) & 0.81 & 20.8 & 0.68 & 2.88 & 6.03 & 238.2 & 0.80 & 0.59 & 31 \\
        \hline
      \end{tabular}}
      \vspace{1em}
    \caption{The impact of Classifier-Free Guidance (CFG) on generation metrics. The tokenizer is DetailFlow-32 with \(N=256\) latent tokens. Reported evaluation metrics include gFID, sFID, Inception Score(IS), Precision (Pre) and Recall (Rec).}
    \label{tab:abla:cfg}
    \centering
    \setlength{\tabcolsep}{1.7mm}{
    \begin{tabular}{c|ccccc}
        \hline
        CFG & gFID\(\downarrow\) & sFID\(\downarrow\) & IS\(\uparrow\) & Pre\(\uparrow\) & Rec\(\uparrow\) \\
        \hline
        1.4 & 2.79 & 5.87 & 229.1 & 0.80 & 0.60 \\
        1.5 & 2.75 & 5.77 & 250.8 & 0.81 & 0.58 \\
        1.6 & 2.81 & 5.66 & 268.1 & 0.82 & 0.57 \\
        1.7 & 3.02 & 5.61 & 283.2 & 0.83 & 0.56 \\
        \hline
      \end{tabular}}
  \end{table}

Table~\ref{tab:abla:prob} reports an ablation study on the impact of different probabilities of coarse-to-fine training on image generation quality. The results show that even a 20\% probability of degraded-resolution training enables the tokenizer to learn a hierarchical token structure effectively. However, increasing this probability further shifts training focus toward reconstructing downsampled images with partial tokens, which hinders the model’s capacity to learn full-sequence representations. These findings highlight the importance of balancing coarse and fine training to optimize efficiency under limited computational resources.

\textbf{Impact of Group Size on Parallel Token Prediction.} DetailFlow employs a parallel decoding strategy that predicts \(g\) tokens simultaneously to accelerate inference by reducing the number of decoding steps. Table~\ref{tab:abla:g} analyzes the impact of varying group sizes \(g\) on image reconstruction and generation, with the total token length fixed at \(N=256\). Results indicate that reconstruction quality remains largely unaffected by changes in \(g\), likely because the total number of tokens is constant. However, image generation quality shows a slight decline of 0.13 gFID as \(g\) increases, potentially due to higher sampling noise from parallel prediction. Despite this, the degradation is minimal, suggesting that the self-correction training effectively counteracts sampling errors. This allows for increased parallelism and faster inference with only a marginal loss in image quality.

\textbf{Impact of Classifier-Free Guidance.} Table~\ref{tab:abla:cfg} presents the impact of varying the Classifier-Free Guidance (CFG) scale on generation quality using DetailFlow-32 as the tokenizer. It can be observed that an appropriate CFG value 1.5 can effectively balance generation quality and diversity.

\subsubsection{State-of-the-art image generation}

Table~\ref{tab:sota:detail} presents a detailed comparison between our proposed method DetailFlow and state-of-the-art approaches under similar AR model sizes. The results demonstrate that our model, DetailFlow-16, achieves a higher gFID score and significantly outperforms existing methods in terms of Recall metric, while attaining comparable performance in Precision metric. These findings indicate that DetailFlow-16 is capable of generating higher-quality images using fewer tokens and at a faster inference speed.

Furthermore, as the token count increases to 256 and 512, the performance of DetailFlow continues to improve. In particular, DetailFlow-32 strikes a strong balance across image quality, diversity, and generation speed. This suggests that, with a comparable or lower training cost (in terms of token number) and inference cost, DetailFlow consistently outperforms existing models, highlighting its efficiency and effectiveness in autoregressive image generation.

\subsection{Future Work}

\begin{table}[t]
    \caption{Comparison of AR methods for class-conditional image generation on ImageNet-1k at \(256 \times 256\) resolution under similar AR model size. Models marked with \(^\dag\) leverage additional training data beyond ImageNet. The \(^\diamond\) symbol denotes models that generate images at \(384 \times 384\) resolution, which are subsequently downsampled to \(256 \times 256\) for evaluation. O.-MAGVIT2 means Open-MAGVIT2. Reported evaluation metrics include rFID, PSNR, gFID, Inception Score(IS), Precision (Pre) and Recall (Rec).}
    \label{tab:sota:detail}
    \centering
    \setlength{\tabcolsep}{0.2mm}{
    \begin{tabular}{l|cc|ccccccccc}
        \hline
        Method & rFID\(\downarrow\) & PSNR\(\uparrow\) & Generator & gFID\(\downarrow\) & IS\(\uparrow\) & Pre\(\uparrow\) & Rec\(\uparrow\) & Param. & \#Tokens & Step & Time(s) \\
        \hline
        LlamaGen\cite{sun2024autoregressive} & 2.19 & 20.79 & GPT-L & 3.81 & 248.3 & 0.83 & 0.52 & 343M & 256 & 256 & 12.58 \\
        O.-MAGVIT2\cite{luo2024open} & 1.17 & 22.64 & AR-B & 3.08 & 258.3 & 0.85 & 0.51 & 343M & 256 & 256 & - \\
        PAR\cite{wang2024parallelized} & 0.94 & - & PAR-L-4 & 3.76\(^\diamond\) & 218.9\(^\diamond\) & 0.84\(^\diamond\) & 0.50\(^\diamond\) & 343M & 576 & 147 & 3.38 \\
        VAR\(^\dag\)\cite{tian2024visual} & 0.9 & - & VAR-d16 & 3.30 & 274.4 & 0.84 & 0.51 & 310M & 680 & 10 & 0.15 \\ 
        FlexVAR\(^\dag\)\cite{jiao2025flexvar} & - & - & VAR-d16 & 3.05 & 291.3 & 0.83 & 0.52 & 310M & 680 & 10 & 0.15 \\ 
        \rowcolor{gray!10}
        \textbf{DetailFlow-16} & 1.22 & 19.39 & GPT-L & 2.96 & 221.4 & 0.82 & 0.57 & 326M & 16*8 & 23 & 0.08 \\
        \rowcolor{gray!10}
        \textbf{DetailFlow-32} & 0.80 & 20.80 & GPT-L & 2.75 & 250.8 & 0.81 & 0.58 & 326M & 32*8 & 39 & 0.16 \\
        \rowcolor{gray!10}
        \textbf{DetailFlow-64} & 0.55 & 22.49 & GPT-L & 2.62 & 245.3 & 0.80 & 0.60 & 326M & 64*8 & 71 & 0.38 \\
        \hline
      \end{tabular}}
  \end{table}

To ensure fair comparison with existing methods, the tokenizer in our experiments is trained on square images with equal height and width. However, our proposed framework, DetailFlow, is not inherently restricted to this setting. Both its encoder and decoder adopt the SigLIP2-NaFlex architecture, which natively supports inputs of arbitrary resolution and aspect ratio. By resizing positional encodings to match input dimensions, the model remains compatible with non-square images. Under this design, the implicit, learnable 1D latent tokens effectively represent images of any resolution or aspect ratio.

To enable autoregressive (AR) models to generate images with a specified aspect ratio, it is essential to condition the model during both training and inference. This can be achieved by incorporating aspect ratio information via natural language prompts or special tokens that encode the target ratio. Given the relationship between image resolution and token count \(r_n = \sqrt{hw} = \mathcal{R}(n)\), the model is guided to predict a specific number of latent tokens corresponding to the desired resolution and aspect ratio.

\subsection{Limitations}

DetailFlow achieves efficient token compression by embedding 2D image information into a 1D coarse-to-fine token sequence using a query-token-driven tokenizer. However, this design introduces limitations, particularly in high-resolution image reconstruction. Capturing fine-grained visual details often requires several thousand latent tokens, substantially increasing the tokenizer's computational cost during training.

In contrast, conventional 2D tokenizers, with spatially consistent strategies, are trainable on low-resolution images and generalize effectively to higher resolutions. 1D tokenizer lacks this scalability, making it less efficient in high-resolution settings.

To mitigate the high training cost, a progressive training strategy proves effective. Since both the encoder and decoder support variable input resolutions, training can begin with low-resolution images and less latent tokens to establish robust encoding and decoding. The model is then fine-tuned on high-resolution data, enabling adaptation to finer spatial details without retraining from scratch. This approach enhances training efficiency while preserving the model’s flexibility across resolutions.

\subsection{More Visual Results}

\begin{figure}
    \begin{center}
        \includegraphics[width=0.99\linewidth]{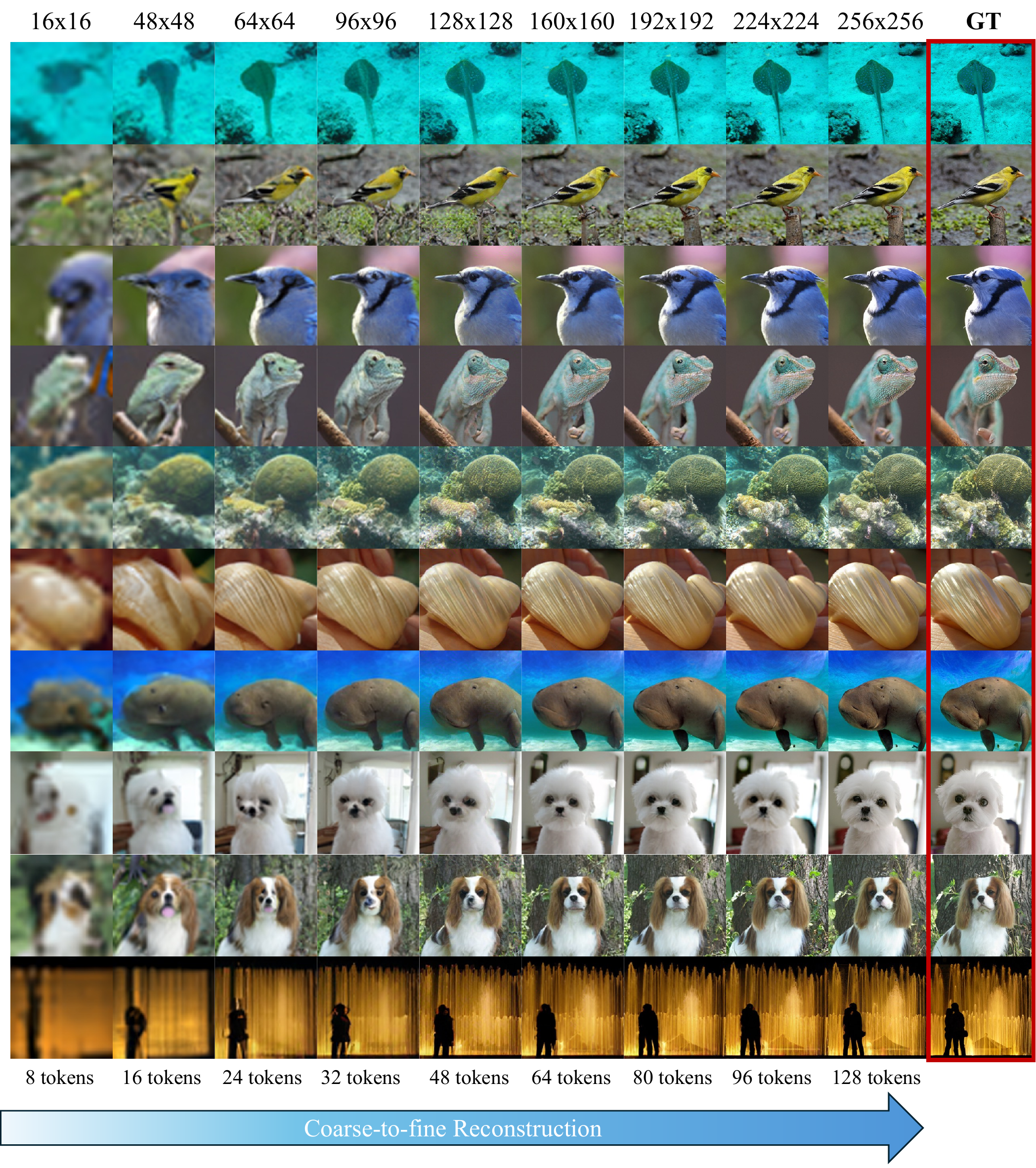}
     \end{center}
    \caption{Progressive reconstruction results from DetailFlow-16. Our method encodes 2D image content into a coarse-to-fine 1D token sequence, with early tokens capturing global structure and later tokens introducing details necessary for both finer texture and higher resolution. As the number of tokens increases, the reconstructed images exhibit progressive improvements in both detail and resolution. The last column is the original Ground Truth (GT) image.}
    \label{fig:demo:vq_case}
  \end{figure}

\begin{figure}
    \begin{center}
        \includegraphics[width=0.99\linewidth]{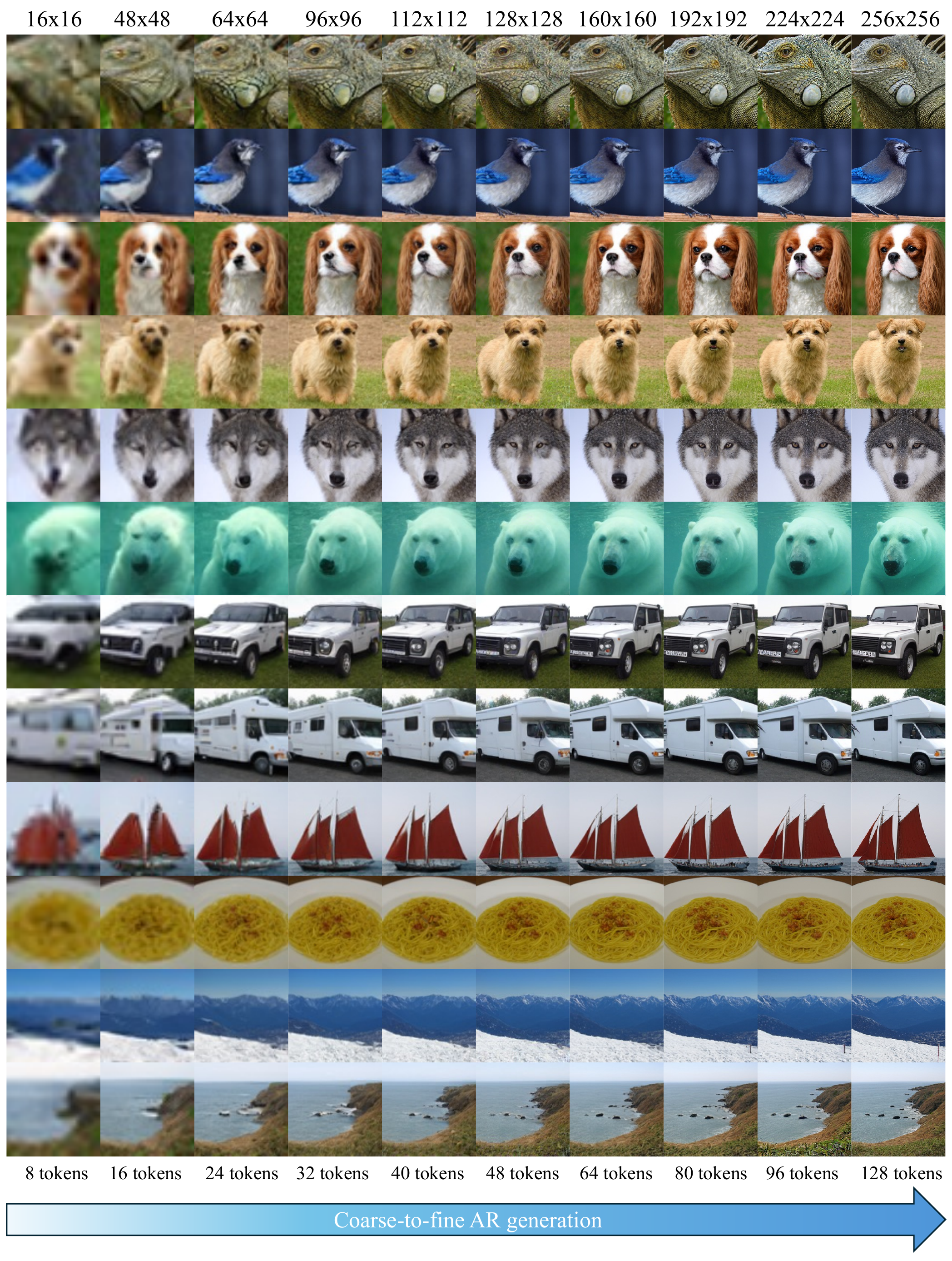}
     \end{center}
    \caption{Progressive generation results from DetailFlow-16. Our proposed 1D tokenizer encodes tokens with an inherent semantic ordering, where each subsequent token contributes additional high-resolution information. The sequences illustrate how image resolution and inferred 1D tokens incrementally increase from left to right.}
    \label{fig:demo:gen_case}
  \end{figure}

Fig.~\ref{fig:demo:vq_case} and Fig.~\ref{fig:demo:gen_case} present reconstruction and generation examples produced by DetaiFlow across various categories. As the number of tokens increases, both the detail and resolution of the images improve accordingly, illustrating a clear coarse-to-fine progression. This observation highlights the effectiveness of our coarse-to-fine training strategy, which imposes a structured semantic order on the token sequence that aligns well with the autoregressive prediction paradigm.


\end{document}